\documentclass[runningheads]{llncs}

\makeatletter
\def\thanks#1{\protected@xdef\@thanks{\@thanks
        \protect\footnotetext{#1}}}
\makeatother

\usepackage{eccv}

\usepackage{eccvabbrv}

\usepackage{graphicx}
\usepackage{booktabs}

\usepackage{makecell}
\usepackage{multirow}
\usepackage{colortbl}
\usepackage{array}
\usepackage{pifont}
\usepackage{wrapfig}
\usepackage{arydshln}

\usepackage[accsupp]{axessibility}  % Improves PDF readability for those with disabilities.

\definecolor{cvprblue}{rgb}{0.21,0.49,0.74}
\usepackage[pagebackref,breaklinks,colorlinks,citecolor=cvprblue]{hyperref}

\definecolor{lightblue}{RGB}{173, 216, 230}
\definecolor{mediumblue}{RGB}{100, 149, 237}
\definecolor{lightpink}{RGB}{255, 182, 193}
\definecolor{lightgreen}{RGB}{144, 238, 144}
\definecolor{mediumgreen}{RGB}{60, 179, 113}
\def\eg{{\em e.g.}}
\def\ie{{\em i.e.}}

\definecolor{mygray}{gray}{0.9}
\definecolor{highlight}{RGB}{238,250,215}
\newcommand{\improve}[1]{{\textcolor[RGB]{61,145,64}{#1}}}

\definecolor{gxred}{RGB}{255, 0, 0}

\usepackage{orcidlink}

\begin{document}

% ---------------------------------------------------------------
% TODO REVIEW: Replace with your title
\title{Towards Long-Form \\ Spatio-Temporal  Video Grounding} 

% TODO REVIEW: If the paper title is too long for the running head, you can set
% an abbreviated paper title here. If not, comment out.
\titlerunning{Towards Long-Form Spatio-Temporal Video Grounding}

% TODO FINAL: Replace with your author list. 
% Include the authors' OCRID for the camera-ready version, if at all possible.
\author{Xin Gu\inst{1} \and
Bing Fan\inst{2} \and Jiali Yao\inst{5} \and Zhipeng Zhang\inst{3} \and Yan Huang\inst{2} \and Cheng Han\inst{4} \\ Heng Fan\inst{2\dag} \and Libo Zhang\inst{5\dag}
\thanks{$^{\dag}$Equal advising and corresponding authors}
}

% TODO FINAL: Replace with an abbreviated list of authors.
\authorrunning{X.~Gu et al.}
% First names are abbreviated in the running head.
% If there are more than two authors, 'et al.' is used.

% TODO FINAL: Replace with your institution list.
\institute{$^{1}$University of Chinese Academy of Sciences\;
$^{2}$UNT \;
$^{3}$Shanghai Jiao Tong University \\
$^{4}$UMKC \; 
$^{5}$Institute of Software Chinese Academy of Sciences}

\maketitle

\begin{abstract}
  In real scenarios, the videos can span several minutes or even hours, yet existing research on spatio-temporal video grounding (STVG), given a textual query, mainly focuses on localizing the target from a video of tens of seconds, typically \emph{less than} one minute, hindering its applications. In this paper, we explore \textbf{L}ong-\textbf{F}orm \textbf{STVG} (\textbf{LF-STVG}), which aims to locate the target from long-term videos. In LF-STVG, long-term videos encompass a much longer temporal span and more irrelevant information, making it challenging for current short-form STVG approaches that process all the frames at once. Addressing these, we propose a novel \textbf{A}uto\textbf{R}egressive \textbf{T}ransformer architecture for LF-\textbf{STVG}, dubbed \textbf{ART-STVG}. Unlike current STVG methods requiring seeing the entire video sequence to make a full prediction at once, our ART-STVG regards the video as a streaming input and processes its frames sequentially, making it capable of easily handling the long videos. To capture spatio-temporal context in ART-STVG, spatial and temporal memory banks are developed and applied to decoders of ART-STVG. Considering that memories at different moments are not always relevant for localizing the target in current frame, we introduce simple yet effective memory selective strategies that enable the more relevant information for decoders, greatly improving the performance. Moreover, rather than parallelizing spatial and temporal localization as done in existing approaches, we introduce a novel cascaded spatio-temporal design that connects spatial decoder to temporal decoder during grounding, which allows ART-STVG to leverage more fine-grained target information to assist with complicated temporal localization in complex long videos, further boosting performance. On the newly extended datasets for LF-STVG, ART-STVG largely outperforms current approaches, while showing competitive results on Short-Form STVG. Our code is at: \url{https://github.com/HengLan/ART-STVG}.

\end{abstract}

\section{Introduction}
\label{intro}

Spatio-temporal video grounding (\textbf{STVG}) aims to localize a target of interest in \emph{space} and \emph{time} from an untrimmed video given a \emph{free-form} textual query~\cite{STGRN}. As a multimodal task, it needs to accurately comprehend the spatio-temporal content of a video and make connections with the given textual query for target localization. Owing to its pivotal role in multimodal video understanding, STVG has recently attracted extensive attention (\eg,~\cite{STGRN,STCAT,su2021stvgbert,hcstvg,TubeDETR,zhang2020object,csdvl,cgstvg,wasim2024videogrounding,gu2025knowing}).

\begin{figure}[!t]
    \centering
    \includegraphics[width=1\linewidth]{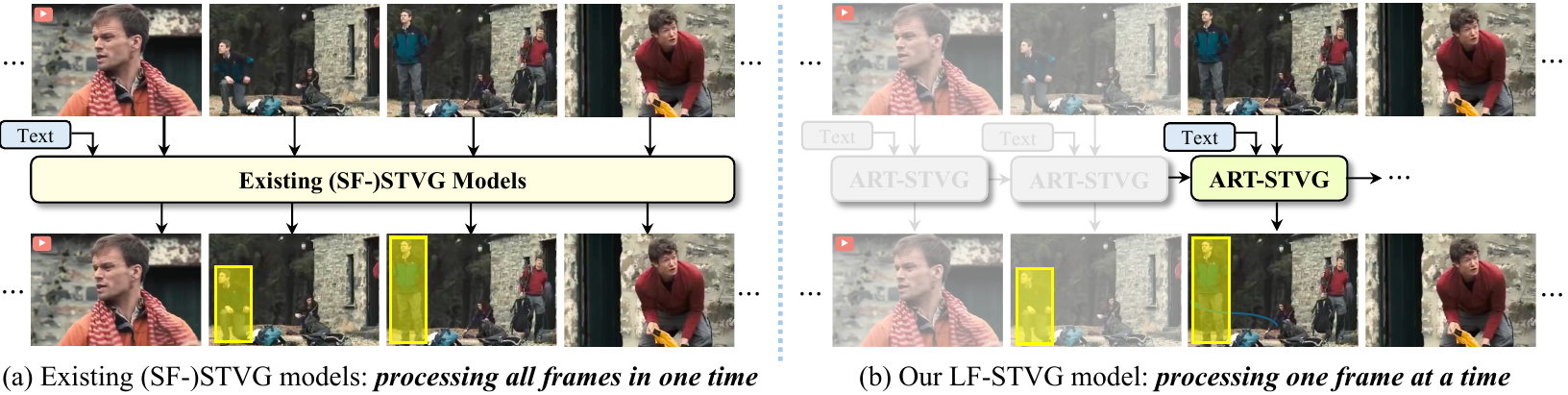}
    \caption{Comparison of existing STVG approaches (\eg,~\cite{TubeDETR,cgstvg,STCAT,csdvl,wasim2024videogrounding,gu2025knowing}) that see the entire video to make a full prediction at once in (a) and ART-STVG that processes frames one at a time and is more suitable for handling long videos in LF-STVG in (b).}\vspace{-1mm}
    \label{fig:task_comparison}
    \vspace{-5mm}
\end{figure}

Despite advancements, current research mainly focuses on locating the target within a short-term video lasting tens of seconds, typically \emph{less than} one minute. For instance, the average video length of existing popular benchmarks HCSTVG-v1/-v2~\cite{hcstvg} and VidSTG~\cite{STGRN} is $20$ and $35$ seconds, respectively. Nevertheless, in real-world applications, such as video retrieval and visual surveillance, the videos can span several \emph{minutes} or even \emph{hours}, resulting in a large \emph{gap} between current research (focusing on target localization within \emph{short-term} videos) and practical applications (the need of target localization in \emph{long-term} videos). To mitigate this gap, in this paper, we explore \textbf{L}ong-\textbf{F}orm \textbf{STVG} (\textbf{LF-STVG}), which localizes the target of interest from \emph{long-term} videos given a textual query.

\setlength{\columnsep}{10pt}
\setlength\intextsep{0pt}
\begin{wrapfigure}{r}{0.5\textwidth}
\centering

\includegraphics[width=0.5\textwidth]{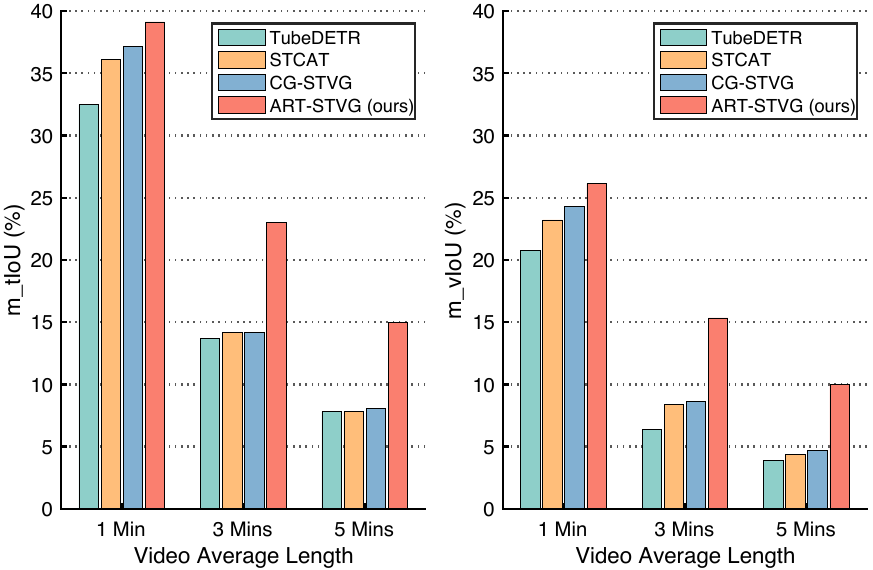}\vspace{-1mm}
\caption{Comparison on the datasets with minute-level videos. Existing STVG methods severely suffer localization in long sequences, and ART-STVG significantly surpasses these models. Moreover, the longer the video is, the larger the improvement of ART-STVG over other methods is.}
\label{fig:compare_2}
\end{wrapfigure}
To localize the desired target, existing STVG approaches (\eg, \cite{wasim2024videogrounding,STCAT,csdvl,TubeDETR,cgstvg,gu2025knowing}) process \emph{all} video frames in one time (see Fig.~\ref{fig:task_comparison} (a)), aiming to capture and leverage global context of the entire video for localization. These  approaches have achieved impressive results on current Short-Form STVG (SF-STVG). Nonetheless, as the video length increases, new challenges arise, prompting an \textbf{important question}: \emph{Is processing all video frames at once, as done in existing SF-STVG models, still applicable to LF-STVG?}

Our response is \textbf{negative}! In LF-STVG, the long video often comprises a longer temporal span, which largely increases the complexities of spatio-temporal localization. In addition, the long video commonly contains far more irrelevant information, requiring a model to identify the target event from extensive redundant content. For these reasons, processing all frames of a long video at once, as done in current STVG methods, presents significant challenges in capturing long-term spatio-temporal relationships and handling excessive irrelevant information for accurate localization (see Fig.~\ref{fig:compare_2}). Additionally, it causes computational bottlenecks because of high GPU memory requirements for simultaneous target localization in all frames.

Addressing the aforementioned challenges, we propose a novel \textbf{A}uto\textbf{R}egressive \textbf{T}ransformer method for LF-\textbf{STVG}, dubbed \textbf{ART-STVG}. Specifically, it treats the video as a streaming input, and processes its frames sequentially (see Fig.~\ref{fig:task_comparison} (b)). To capture the crucial spatio-temporal contextual information in the video, we maintain two memory banks, that reserve essential spatio-temporal information from the video, for spatial and temporal decoders in ART-STVG. Since the memories in the banks are \emph{not} equally important to a certain frame, we introduce simple yet effective memory selective strategies to leverage more relevant information in the memory banks for grounding, effectively boosting performance. Compared to existing approaches that require seeing the entire video for prediction, our proposed ART-STVG ingests frames one at a time for prediction, hence naturally processing longer videos and resolving the computational bottleneck faced by current approaches. Furthermore, rather than parallelizing the spatial and temporal localization as is done in existing approaches, we propose a novel cascaded spatio-temporal design which connects spatial decoder to temporal decoder during grounding. By doing this, our ART-STVG can enjoy more fine-grained target information from the spatial decoder to assist with the more complicated temporal localization, further enhancing performance. To our best knowledge, this paper is the \emph{first} to explore the LF-STVG problem, and ART-STVG is the \emph{first} framework attempting to handle LF-STVG.

To verify the effectiveness of our ART-STVG in long-term videos, we extend the validation set of the short-term benchmark HCSTVG-v2~\cite{hcstvg} (the reason for choosing HCSTVG-v2 for the extension is described later).
Specifically, we extend its average video length from $20$ seconds to 1/3/5 minutes, hence it is referred to as LF-STVG-1min/3min/5min. 
We conduct extensive experiments on both long-form and short-form STVG. The results demonstrate that our ART-STVG outperforms all existing approaches on LF-STVG by achieving new state-of-the-art results while showing competitive performance on SF-STVG.

In summary, our \textbf{contributions} are as follows: \ding{171} We explore the LF-STVG problem and introduce a novel memory-augmented autoregressive transformer, dubbed ART-STVG, for LF-STVG; \ding{170} We propose memory selection strategies that allow the selection of relevant crucial spatio-temporal context to enhance target localization; \ding{168} We introduce a cascaded spatio-temporal decoder design to fully leverage the fine-grained information produced by spatial localization to assist in temporal localization; \ding{169} In extensive experiments on both long-term and short-term benchmarks, our ART-STVG achieves excellent performance. 
\vspace{-0.1em}

\section{Related Work}

\textbf{Spatio-temporal video grounding} aims to locate the tube in the video that corresponds to a given textual query. Early methods (\eg,~\cite{2d-tan,pcc,mmn,STGRN,su2021stvgbert}) are predominantly two-stage approaches. These approaches first use a pre-trained object detector~\cite{faster} to generate object proposals and then select the proposals based on the given textual query. Such methods are easily limited by the pre-trained object detector. Recent methods (\eg,~\cite{STCAT,csdvl,talal2023video,cgstvg,gu2025knowing}), inspired by DETR~\cite{dert}, propose one-stage frameworks that directly generate tubes for target localization, performing better than two-stage models.
Nevertheless, both early two-stage and recent one-stage approaches focus on SF-STVG and process the entire video at one time for simultaneous target localization in all frames. \emph{\textbf{Different from}} existing approaches, our ART-STVG is specially designed for LF-STVG. Specifically, ART-STVG treats the video as a streaming input and processes its frames sequentially with an autoregressive framework, thus making it more suitable for handling long-term videos. Besides, we design memory selection strategies specifically for this streaming framework to enhance localization.

\vspace{0.3em}
\noindent
\textbf{Long-term video understanding} has received significant attention recently in various tasks such as action detection (\eg,~\cite{varol2017long,cheng2022tallformer}), video captioning (\eg,~\cite{li2022long,islam2024video}), and video question answering (\eg,~\cite{song2024moviechat,maaz2023video,cheng2024videollama,he2024ma}). 
The major challenge of long video understanding is that capturing complex spatio-temporal dependencies over long durations requires a high computational cost.
To address this, early methods (\eg,~\cite{donahue2015long,wu2021towards,girdhar2017actionvlad,hussein2019timeception}) model pre-extracted video features without jointly training the backbone. Recent works (\eg,~\cite{wu2019adaframe,bai2023qwen,zhang2024long}) present efficient strategies to process more frames simultaneously, while others (\eg,~\cite{wu2022memvit,he2024ma,qian2025streaming,wang2024videollamb,fan2019heterogeneous}) construct streamlined transformers and integrate memory banks for effective long video understanding.
\emph{\textbf{Different from}} these works, we focus on LF-STVG. Particularly, \textbf{\emph{unlike}} video query answering~\cite{song2024moviechat,he2024ma} in which the memory is global context information for the entire video, the memory in our ART-STVG is text-guided spatial instance and temporal event boundary cues, specifically designed for LF-STVG. In addition, for memory selection, our method
merges spatial memories using text as guidance and temporal memories using event boundary cues from videos, while the other methods~\cite{song2024moviechat,he2024ma} do not have these mechanisms because they aim at different tasks.

\vspace{0.3em}
\noindent
\textbf{Autoregressive architecture} has been extensively studied and applied in various domains. Early autoregressive models include recurrent neural networks (RNN)~\cite{medsker2001recurrent}, long-short-term memory networks (LSTM)~\cite{graves2012long,hochreiter1997long}, and gated recurrent units (GRU)~\cite{chung2014empirical}. Recently, autoregressive transformer models (\eg,~\cite{vaswani2017attention,katharopoulos2020transformers,touvron2023llama, liu2024visual,ren2024timechat,lin2023video}) by applying self-attention mechanism have further advanced the field by enabling serial computation and capturing the long-range dependencies. \emph{\textbf{Different from}} the aforementioned methods, in this paper we propose an autoregressive transformer architecture for LF-STVG.
\vspace{-0.1em}

\section{The Proposed Approach}

\begin{figure}[!t]
    \centering
    \includegraphics[width=0.96\linewidth]{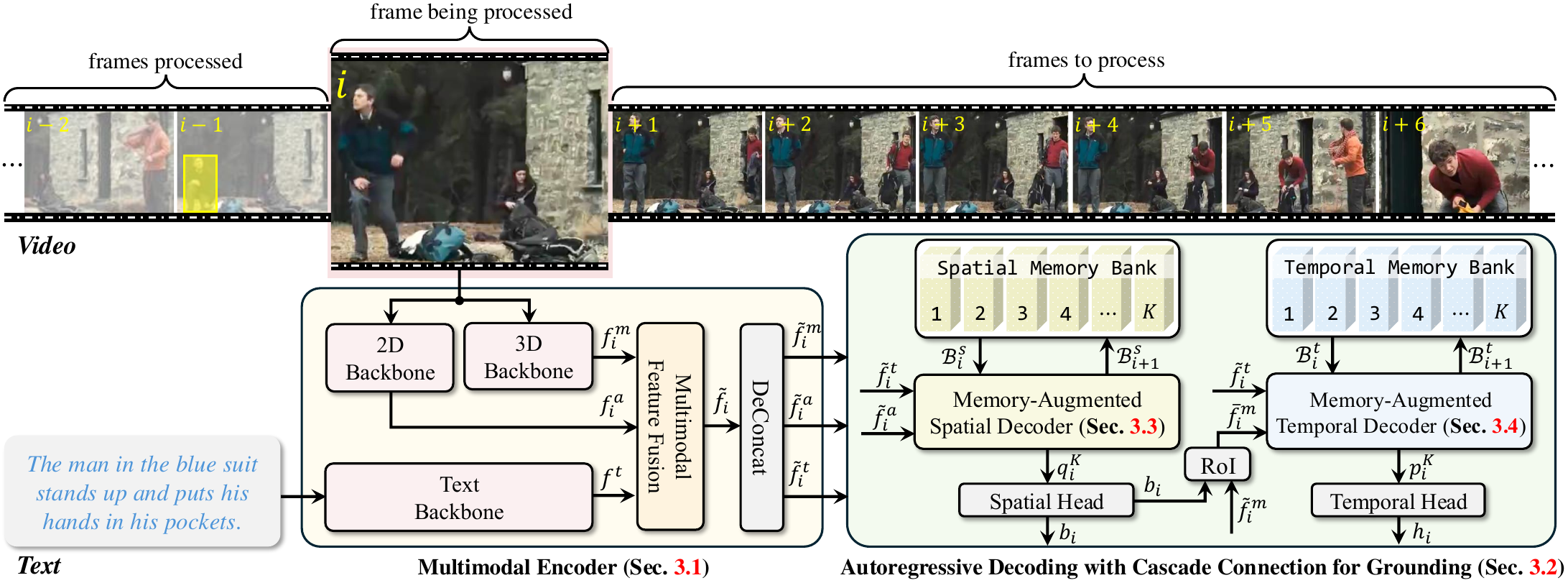}\vspace{-1mm}
    \caption{Overview of ART-STVG, comprising a multimodal encoder and an autoregressive decoder with cascade connection for target localization frame by frame.}
    \label{fig:framework}
    \vspace{-4.5mm}
\end{figure}

\textbf{Overview.} We propose ART-STVG, a memory-augmented autoregressive transformer for LF-STVG. As shown in Fig.~\ref{fig:framework}, the framework begins with a multimodal encoder (Sec.~\ref{encoder}) that extracts and fuses visual and textual features. Following this, the cascaded spatio-temporal decoder performs autoregressive decoding for grounding (Sec.~\ref{autoregressive}). Specifically, the memory-augmented spatial decoder (Sec.~\ref{sdecoder}) captures the spatial location information of the target, while the memory-augmented temporal decoder (Sec.~\ref{tdecoder}) focuses on learning the temporal location information. 

Since ART-STVG processes frames sequentially, in the following, we take the processing of the $i^{\text{th}}$ frame as an example to illustrate our ART-STVG.

\subsection{Multimodal Encoder}
\label{encoder}

Given the $i^{\text{th}}$ frame of a video and the corresponding textual query, the multimodal encoder generates a multimodal feature, which is sent to the autoregressive decoder for localization. The multimodal encoder comprises feature extraction and feature fusion, as described below.

\vspace{0.3em}
\noindent
\textbf{Feature Extraction.} For the $i^{th}$ frame of the video sequence, we separately extract its 2D appearance and 3D motion features to leverage rich static and dynamic cues. Specifically, the appearance feature is extracted using ResNet-101~\cite{resnet}, and the motion feature is extracted through VidSwin~\cite{vidswin}. Please \textbf{\emph{note}}, when applying VidSwin to extract motion features, the $(i-1)^{\text{th}}$ frame is also used as input. The appearance feature of frame $i$ is denoted as $f_i^a \in \mathbb{R}^{H \times W \times C_a}$, where $H$, $W$, and $C_a$ are height, width, and channel dimensions. Similarly, the motion feature is represented as $f_i^m \in \mathbb{R}^{H \times W \times C_m}$ with $C_m$ the channel dimension.
For the text, we first tokenize it to a word sequence, and then apply RoBERTa~\cite{roberta} to extract its feature $f^t \in \mathbb{R}^{N_t \times C_t}$, where $N_t$ is the text feature length and $C_t$ represents the channel dimension.

\vspace{0.3em}
\noindent
\textbf{Feature Fusion.} Different modalities typically contain complementary information. Therefore, we fuse the appearance feature $f_i^a$ and motion feature $f_i^m$ of the $i^{\text{th}}$ video frame with the textual feature $f^t$ to generate a multimodal feature of the $i^{\text{th}}$ frame. Specifically, we first project them to the same channel dimension $C$, and then concatenate them to produce the multimodal feature $f^{'}_i$, as follows, 
\begin{equation}\nonumber
\setlength{\abovedisplayskip}{6pt}
\setlength{\belowdisplayskip}{6pt}
    f^{'}_i = [\underbrace{f^a_{i_1}, f^a_{i_2}, ..., f^a_{i_{H \times W}}}_{\text{appearance feature $f^a_i$}},\underbrace{f^m_{i_1}, f^m_{i_2}, ..., f^m_{i_{H \times W}}}_{\text{motion feature $f^m_i$}}, \underbrace{f_{1}^{t}, f_{2}^{t}, ..., f_{N_t}^{t}}_{\text{textual feature $f^t$}}]
\end{equation}
Afterwards, we apply a self-attention encoder~\cite{vaswani2017attention} to further fuse the multimodal features, as follows,
\begin{equation}\nonumber
\setlength{\abovedisplayskip}{6pt}
\setlength{\belowdisplayskip}{6pt}
    \tilde{f}_i = \mathtt{SelfAttEncoder}(f^{'}_i+\mathcal{E}_{pos}+\mathcal{E}_{typ})
\end{equation}
where $\mathcal{E}_{pos}$ and $\mathcal{E}_{typ}$ denote position and type embeddings, and $\mathtt{SelfAttEncoder}(\cdot)$ is the self-attention encoder~\cite{vaswani2017attention} with $N$ ($N$=6) self-attention encoder blocks. 

After obtaining the $\tilde{f}_i$, we deconcatenate it to generate enhanced appearance, motion, and textual features $\tilde{f}_i^{a}$, $\tilde{f}_i^{m}$, and $\tilde{f}_i^{t}$ via $[\tilde{f}_i^{a},\tilde{f}_i^{m},\tilde{f}_i^{t}]=\mathtt{DeConcat}(\tilde{f}_i)$ and apply them in the subsequent decoder for target localization.

\subsection{Autoregressive Decoding for Grounding}
\label{autoregressive}

ART-STVG autoregressively decodes video frames to sequentially predict spatio-temporal positions of target. As displayed in Fig.~\ref{fig:framework}, the decoding process of ART-STVG contains two parts, including spatial grounding and temporal grounding via two decoders. The former is responsible for predicting the spatial location of the target, while the latter generates the temporal location of the target event. To capture spatio-temporal context in ART-STVG, spatial and temporal memory banks storing historical information, with effective memory selection strategies, are developed and applied in the grounding process, largely enhancing  performance. Besides, \emph{rather than paralleling} the spatial and temporal grounding as done in current approaches, we propose a novel \emph{\textbf{cascaded}} design to connect spatial and temporal grounding in ART-STVG (see decoding part in Fig.~\ref{fig:framework}). Such cascaded spatio-temporal architecture allows ART-STVG to employ more fine-grained target cues, provided by spatial grounding, to assist with temporal localization in complex long videos, further improving ART-STVG for LF-STVG. 

\vspace{0.35em}
\noindent
\textbf{Spatial Grounding.} In ART-STVG, the spatial grounding is achieved by learning a spatial query via iterative interaction with the multimodal feature. Let $q^0_{i}$ denote the initial spatial query in the $i^{\text{th}}$ frame and $\mathcal{B}^s_{i}$ is the spatial memory bank at this moment. Given the appearance feature $\tilde{f}^a_i$ and textual feature $\tilde{f}^t_i$ from $\tilde{f}_i$ in frame $i$, the interaction of the spatial query with multimodal feature is achieved as follows,
\begin{equation}\nonumber
\setlength{\abovedisplayskip}{6pt}
\setlength{\belowdisplayskip}{6pt}
    q_{i}^K, \mathcal{B}^s_{i+1} = \mathtt{MA\text{-}SpatialDecoder}(q_i^0, \mathcal{B}^s_i, [\tilde{f}^a_i, \tilde{f}^t_i])
\end{equation}
where $\mathtt{MA\text{-}SpatialDecoder}(\cdot)$ is the memory-augmented spatial decoder with $K$ spatial decoder blocks (described in Sec.~\ref{sdecoder}). It is worth \textbf{\emph{noting}} that, the spatial memory bank $\mathcal{B}^s_i$ contains $K$ (\ie, the number of decoder blocks) partitions, with each partition corresponding to a spatial decoder block. $q_{i}^K$ represents the final spatial query feature after $K$ decoder blocks, and $\mathcal{B}^s_{i+1}$ the new memory bank updated with spatial information from frame $i$ (see Sec.~\ref{sdecoder}). After this, a spatial head, containing an MLP module, is used to predict the final object box $b_i$, as follows,
\begin{equation}\nonumber
\setlength{\abovedisplayskip}{6pt}
\setlength{\belowdisplayskip}{6pt}
    b_i = \mathtt{SpatialHead}(q_{i}^K)
\end{equation}
where $b_i\in \mathbb{R}^{4}$ represents the central position, width, and height of the predicted target box in the $i^{\text{th}}$ frame.

\vspace{0.35em}
\noindent
\textbf{Temporal Grounding.} Similar to spatial grounding, we learn a temporal query for temporal grounding by interacting with the multimodal feature. To exploit fine-grained spatial target information to assist with temporal grounding, we design a cascade architecture. Specifically, with target box $b_i$ from spatial grounding, we first extract fine-grained target motion feature $\bar{f}^m_i \in \mathbb{R}^{1 \times 1 \times \mathcal{C}}$ using RoI pooling~\cite{faster}, as follows,
\begin{equation}\nonumber
\setlength{\abovedisplayskip}{6pt}
\setlength{\belowdisplayskip}{6pt}
    \bar{f}^m_i = \mathtt{RoI}(\tilde{f}^m_i ,b_i)
\end{equation}
Compared to $\tilde{f}^m_i$, $\bar{f}^m_i$ focuses more on target region and thus benefits localization. 

After this, we interact the temporal query with multimodal feature. Let $p_i^0$ denote the initial temporal query in frame $i$ and $\mathcal{B}^t_i$ the temporal memory bank at this moment. With fine-grained motion feature $\bar{f}^m_i$ and textual feature $\tilde{f}^t_i$, the interaction of temporal query and multimodal feature is performed as follows,
\begin{equation}\nonumber
\setlength{\abovedisplayskip}{6pt}
\setlength{\belowdisplayskip}{6pt}
    p_i^K, \mathcal{B}^t_{i+1} = \mathtt{MA\text{-}TemporalDecoder}(p_i^0, \mathcal{B}^t_i, [\bar{f}^m_i, \tilde{f}^t_i])
\end{equation}
where $\mathtt{MA\text{-}TemporalDecoder}(\cdot)$ is the memory-augmented temporal decoder with $K$ temporal decoder blocks (described in Sec.~\ref{tdecoder}). Similar to $\mathcal{B}^s_i$, the temporal  memory bank $\mathcal{B}^t_i$ also comprises $K$ partitions, with each corresponding to a temporal decoder block. $p_i^K$ is the final temporal query feature after the decoder, and $\mathcal{B}^t_{i+1}$ the new memory bank updated with temporal information in frame $i$ (see Sec.~\ref{tdecoder}). After this, a temporal head implemented with an MLP module is adopted for temporal localization in frame $i$, as follows,
\begin{equation}\nonumber
\setlength{\abovedisplayskip}{6pt}
\setlength{\belowdisplayskip}{6pt}
    h_i = \mathtt{TemporalHead}(p_i^K)
\end{equation}
where $h_i\in \mathbb{R}^{2}$ contains event start and end probabilities $h_i^s$ and $h_i^e$ for frame $i$.

By sequentially performing spatial and temporal grounding, we achieve target localization in each frame $i$, and meanwhile adopt information in frame $i$ to update the memory banks for target localization in frame ($i$+1). 

\begin{figure*}[!t]
    \centering
    \includegraphics[width=0.95\linewidth]{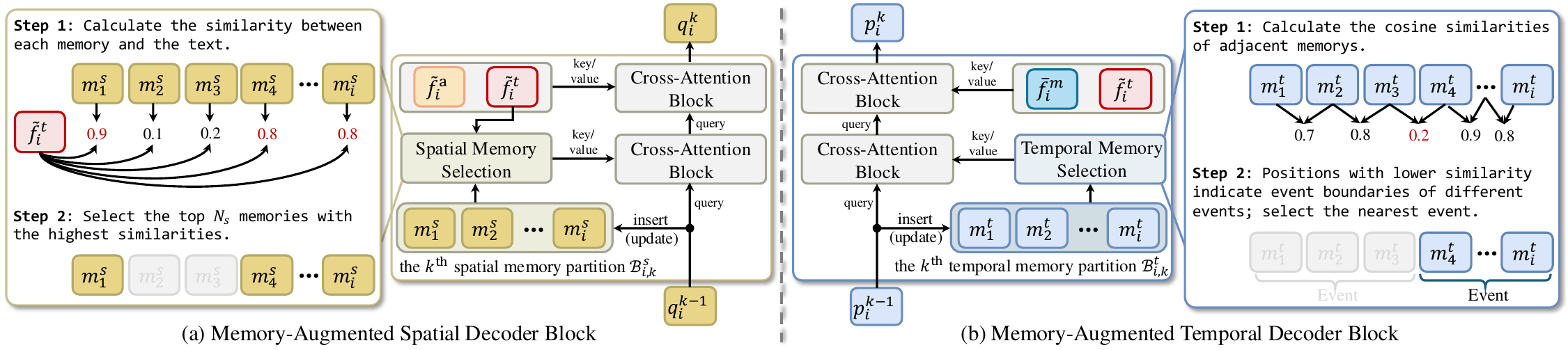}
    \caption{\small{Illustration of the architectures for memory-augmented spatial decoder block in (a) and temporal decoder block in (b).}}
    \label{fig:stdecoder}
    \vspace{-5mm}
\end{figure*}

\subsection{Memory-Augmented Spatial Decoder}
\label{sdecoder}
We propose a memory-augmented spatial decoder, guided by spatial memory from the spatial memory bank, to learn the target spatial position from the multimodal feature. Specifically, the memory-augmented spatial decoder comprises $K$ decoder blocks in a cascade for spatial grounding. As shown in Fig.~\ref{fig:stdecoder} (a), each spatial decoder block corresponds to a partition in the spatial memory and contains two cross-attention blocks~\cite{vaswani2017attention}. Concretely, in the $k^{\text{th}}$ ($1 \le k \le K$) spatial decoder block, given the appearance feature $\tilde{f}^a_i$, the textual feature $\tilde{f}^t_i$, and the spatial query $q_{i}^{k-1}$ ($q_i^{0}$ initialized by zeros) of the $i^{\text{th}}$ frame, we first perform memory selection and then apply the selected memory to enhance the spatial query feature in spatial decoding.

\vspace{0.3em}
\noindent
\textbf{Spatial Memory Selection.} Since the spatial query contains crucial target information, we first insert the spatial query $q_{i}^{k-1}$ into the $k^{\text{th}}$ partition of spatial memory bank $\mathcal{B}^s_{i,k}$ corresponding to the $k^{\text{th}}$ decoder block. Please \emph{\textbf{note}} that, this insertion procedure also completes the update of each partition $\mathcal{B}^s_{i,k}$ in $\mathcal{B}^s_{i}$ to $\mathcal{B}^s_{i+1,k}$ in $\mathcal{B}^s_{i+1}$. Or in other words, we update the memory bank by simply adding the query as a new memory, without removing any existing memories. 

After this, we perform memory selection from $\mathcal{B}^s_{i+1,k}$ for decoder block $k$. The \emph{motivation} behind this selection is, the memories at different moments are not always relevant for target localization in current frame, and selecting more relevant information in the spatial decoding enables learning better query feature for grounding. Specifically, the selective spatial memory $\mathcal{M}^s_{i,k}$ for block $k$ can be obtained via two steps in memory selection: \emph{first}, we calculate the similarity between each spatial memory and the textual feature; \emph{second} , based on the similarity scores, the top $N_s$ spatial memories with the highest scores are selected to form $\mathcal{M}^s_{i,k}$. Fig.~\ref{fig:stdecoder} (a)-left illustrates this spatial memory selection process.

\begin{figure}[!t]
    \centering
\includegraphics[width=0.95\linewidth]{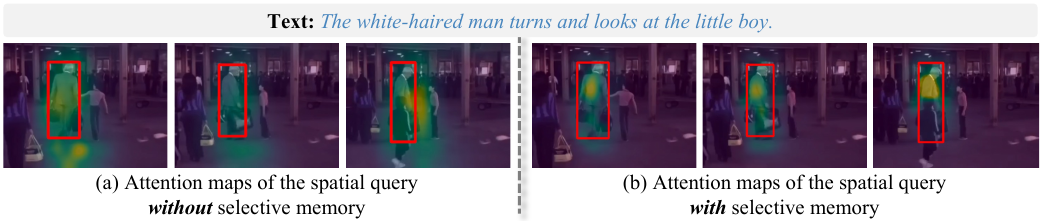}\vspace{-2mm}
    \caption{\small{Comparison of attention maps of the spatial query  \textit{without} and \textit{with} using selective spatial memory. The red box indicates the target object to be located. Through comparison, we observe the use of selective spatial memory helps the model focus more on the target regions, which benefits the final target localization.}}
    \label{fig:spa_att}
    \vspace{-5mm}
\end{figure}

\vspace{0.3em}
\noindent
\textbf{Memory-Augmented Spatial Decoding.} During decoding, we send $q_{i}^{k-1}$ to decoder block $k$ for learning $q_{i}^{k}$. To exploit spatial context, we first interact the query with selective spatial memory through a cross-attention block, as follows,
\begin{equation}\nonumber
\setlength{\abovedisplayskip}{6pt}
\setlength{\belowdisplayskip}{6pt}
    \tilde{q}_{i}^{k-1} = \mathtt{CrossAtt}(q_i^{k-1}, \mathcal{M}^s_{i,k})
\end{equation}
where $\tilde{q}_{i}^{k-1}$ denotes the memory-augmented query feature in decoder block $k$, and $\mathtt{CrossAtt}(\textbf{u},\textbf{v})$ is the cross-attention block~\cite{vaswani2017attention}, with $\textbf{u}$ generating query and $\textbf{v}$ key/value. After this, we further interact $\tilde{q}_{i}^{k-1}$ with the multimodal appearance and textual features for learning $q_{i}^{k}$, as follows,
\begin{equation}\nonumber
\setlength{\abovedisplayskip}{6pt}
\setlength{\belowdisplayskip}{6pt}
    q_{i}^{k} = \mathtt{CrossAtt}(\tilde{q}_{i}^{k-1}, [\tilde{f}^a_i, \tilde{f}^t_i])
\end{equation}
where $q_{i}^{k}$ is the learned query feature, which is sent to the next decoder block for further query feature learning.

In Fig.~\ref{fig:spa_att}, we compare the attention map of the spatial query with and without using selective spatial memory. We can see using selective spatial memory helps the model focus more on target regions for better localization. After $K$ spatial decoder blocks, the spatial query feature $q_i^{K}$ is generated for spatial prediction.

\begin{figure*}[!t]
    \centering
    \includegraphics[width=\linewidth]{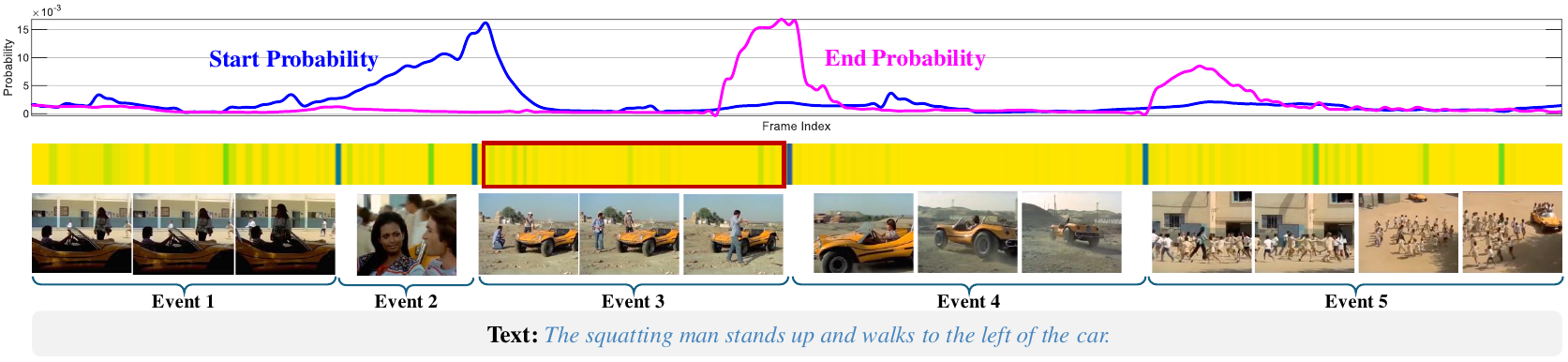}\vspace{-2mm}
    \caption{\small{Illustration of selective temporal memory. In the middle figure, the attention sequence indicates cosine similarities of adjacent memories. The lower similarity (greener color) indicates the potential event boundaries. Besides, in the up figure, we show predicted start and end probabilities, which are accurate to capture the target event. The red box denotes the ground truth corresponding to the text query.}}
    \label{fig:att_seq}
    \vspace{-4mm}
\end{figure*}

\subsection{Memory-Augmented Temporal Decoder}
\label{tdecoder}

The memory-augmented temporal decoder learns target temporal position using temporal memory from temporal memory bank. It has $K$ blocks for temporal grounding, with each corresponding to a temporal memory partition and containing two cross-attention blocks, as in Fig.~\ref{fig:stdecoder} (b). In temporal decoder block $k$ ($1 \le k \le K$), given motion and textual features $\bar{f}^m_i$ and $\tilde{f}^t_i$, and temporal query $p_{i}^{k-1}$ ($p_i^{0}$ initialized by zeros), we first perform temporal memory selection and then apply the selected memory to enhance temporal decoding.

\vspace{0.3em}
\noindent
\textbf{Temporal Memory Selection.} 
The temporal query $p_i^{k-1}$ contains temporal event information. Thus, we first insert it into the $k^{th}$ partition of temporal memory bank $\mathcal{B}^t_{i,k}$ (also updating $\mathcal{B}^t_{i,k}$ to $\mathcal{B}^t_{i+1,k}$). Since long-term videos often contain multiple events, selecting relevant temporal memory related to the current event helps the temporal decoding better locate the event boundaries. To achieve this and obtain selective temporal memory $\mathcal{M}^t_{i,k}$, inspired by TextTiling~\cite{hearst1997text}, we perform two steps in temporal memory section: in the \emph{first} step, we calculate the similarities between the memories of adjacent frames; in the \emph{second} step, points with lower similarities are considered the event boundaries between different events, and we only select the memories corresponding to the event closest to the current frame, as shown in Fig.~\ref{fig:stdecoder} (b)-right.

\vspace{0.3em}
\noindent
\textbf{Memory-Augmented Temporal Decoding.} In decoding, we send the temporal query $p_i^{k-1}$ to temporal decoder block $k$ for learning $p_i^{k}$. To exploit temporal context for enhancing query learning, we first interact the query with the selective temporal memory $\mathcal{M}^t_{i,k}$ by a cross attention block as follows,
\begin{equation}\nonumber
\setlength{\abovedisplayskip}{6pt}
\setlength{\belowdisplayskip}{6pt}
    \tilde{p}_{i}^{k-1} = \mathtt{CrossAtt}(p_i^{k-1}, \mathcal{M}^t_{i,k})
\end{equation}
where $\tilde{p}_{i}^{k-1}$ represents the memory-augmented query feature in decoder block $k$.
After this, we further interact $\tilde{p}_{i}^{k-1}$ with multimodal motion and textual features, as follows,
\begin{equation}\nonumber
\setlength{\abovedisplayskip}{6pt}
\setlength{\belowdisplayskip}{6pt}
    p_{i}^{k} = \mathtt{CrossAtt}(\tilde{p}_{i}^{k-1}, [\bar{f}^m_i, \tilde{f}^t_i])
\end{equation}
where $p_{i}^{k}$ represents the learned query feature, which is sent to the next decoder block for further query feature learning. As shown in Fig.~\ref{fig:att_seq}, the temporal memory selection strategy segments the video into different events and selects the memory closest to current moment, benefiting localization of target event. After $K$ blocks in decoder, the temporal query feature $p_i^{K}$ is adopted for temporal prediction.

\subsection{Optimization}

Given the video containing $N_f$ frames and its textual query, ART-STVG predicts (1) the object boxes $\mathcal{B}=\{b_{i}\}_{i=1}^{N_f}$ in the memory-augmented spatial decoder and event start timestamps $\mathcal{H}_s=\{h_{i}^{s}\}_{i=1}^{N_f}$ and end timestamps $\mathcal{H}_e=\{h_{i}^{e}\}_{i=1}^{N_f}$ in the memory-augmented temporal decoder. During training, with the groundtruth of the bounding box $\mathcal{B}^*$, start timestamps $\mathcal{H}_s^*$ and end timestamps $\mathcal{H}_e^*$, we can calculate the total loss $\mathcal{L}$ af follows,
\begin{equation}\nonumber
\setlength{\abovedisplayskip}{6pt}
\setlength{\belowdisplayskip}{6pt}
        \mathcal{L} =\underbrace{\lambda_k (\mathcal{L}_{{\text{KL}}}(\mathcal{H}_s^*,\mathcal{H}_s) + \mathcal{L}_{{\text{KL}}}(\mathcal{H}_e^*,\mathcal{H}_e))}_{\text{loss of memory-augmented temporal decoder}} + \underbrace{\lambda_l \mathcal{L}_{1}(\mathcal{B}^*,\mathcal{B}) + \lambda_u \mathcal{L}_{{\text{IoU}}}(\mathcal{B}^*,\mathcal{B})}_{\text{ loss of memory-augmented spatial decoder}}
\end{equation}
where $\mathcal{L}_{{\text{KL}}}$, $\mathcal{L}_{1}$ and $\mathcal{L}_{{\text{IoU}}}$ are KL divergence, smooth L1 and IoU losses. $\lambda_k$, $\lambda_l$ and $\lambda_u$ are parameters to balance the training loss.
\vspace{-0.1em}

\section{Experiments}

\textbf{Implementation.} Our ART-STVG is implemented based on PyTorch~\cite{paszke2019pytorch}. We use ResNet-101~\cite{resnet} for appearance feature extraction, VidSwin-tiny~\cite{vidswin} for motion feature extraction, and RoBERTa-base~\cite{roberta} for textual feature extraction. Following previous work~\cite{cgstvg, STCAT,csdvl}, we use pre-trained MDETR~\cite{mdetr} to initialize the appearance and text backbones and the multimodal fusion module. The hidden dimension of the encoder and decoder is $C=256$, with channel dimensions of $C_a = 2048$ for appearance features, $C_m = 768$ for motion features, and $C_t = 768$ for textual features. We sample video frames at FPS of 3.2 and resize each frame to have a short side of $420$. The video frame length during training is $N_f=64$, and the text sequence length is $N_t=30$. During training, we adopt Adam~\cite{kingma2014adam} with an initial learning rate of $1e-5$ for the pre-trained backbone and $1e-4$ for other modules, while keeping the motion backbone frozen. Similar to~\cite{cgstvg, STCAT,csdvl}, the parameters $\lambda_k$, $\lambda_l$ and $\lambda_u$ are empirically set to $10$, $5$, and $3$.

\vspace{0.35em}
\noindent
\textbf{Datasets.} Since there are no available benchmarks available for LF-STVG, we opt to extend HCSTVG-v2~\cite{hcstvg} for creating new datasets for LF-STVG. The \textbf{\emph{reason}} for choosing HCSTVG-v2 only for extension is that it is the \emph{\textbf{only}} benchmark which provides available source videos, thus allowing for extension with longer videos. Specifically, HCSTVG-v2 originally contains 16,000 video-sentence pairs in complex scenes, including 10,131 training samples, 2,000 validation samples, and 4,413 testing samples. Each video lasts 20 seconds and is paired with a natural query sentence averaging 17.25 words. As annotations of the test set are not publicly available, the results are reported on the validation set, as in existing methods~\cite{TubeDETR,csdvl,cgstvg}. For this reason, we extend only the validation set to lengths of 1/3/5 minutes, referred to as LF-STVG-1min/3min/5min, for the evaluation of LF-STVG. The extensions are based on original YouTube videos, not concatenated clips and we extend the HCSTVG-v2 val clips by randomly padding frames on either side. We only constrain total video length, while randomizing extension duration and direction. Three domain-related students manually verify that each grounding tube remains unique and aligned with the textual query. The split and evaluation protocol follow the HCSTVG-v2.
It is worth \textbf{\emph{noting}} that we do not change the training set, and ART-STVG is trained using the same samples as in short-form STVG approaches for a fair comparison.

\vspace{0.35em}
\noindent
\textbf{Metrics.}
Follow previous works~\cite{csdvl,STCAT,cgstvg}, we utilize the commonly used metrics m\_tIoU, m\_vIoU, and vIoU@R for evaluation.
m\_tIoU evaluates the effectiveness of temporal grounding by averaging tIoU scores across the entire test set. m\_vIoU assesses spatial grounding performance by averaging vIoU scores. Additionally, vIoU@R measures performance by determining the proportion of test samples with vIoU scores exceeding a threshold R. For a more detailed explanation of the evaluation metrics used, please refer to previous works~\cite{csdvl,STCAT,cgstvg}.
 
\renewcommand{\arraystretch}{1.1}
\begin{table}[!t]
  \centering
  \setlength{\tabcolsep}{5pt}
  \caption{Comparison to other methods on long-term videos. The best result is highlighted in the \textbf{bold} font.}\vspace{-2mm}
  \resizebox{0.97\textwidth}{!}{
    \begin{tabular}{rcccccccccccc}
   \Xhline{1.2pt}
    & \multicolumn{4}{c}{\textbf{LF-STVG-1min}}      & \multicolumn{4}{c}{\textbf{LF-STVG-3min}}      & \multicolumn{4}{c}{\textbf{LF-STVG-5min}} \\
    
    \cmidrule(lr){2-5} \cmidrule(lr){6-9} \cmidrule(lr){10-13}
        
    & \rotatebox{90}{m\_tIoU} & \rotatebox{90}{m\_vIoU} & \rotatebox{90}{vIoU@0.3} & \rotatebox{90}{vIoU@0.5} & \rotatebox{90}{m\_tIoU} & \rotatebox{90}{m\_vIoU} & \rotatebox{90}{vIoU@0.3} & \rotatebox{90}{vIoU@0.5} & \rotatebox{90}{m\_tIoU} & \rotatebox{90}{m\_vIoU} & \rotatebox{90}{vIoU@0.3} & \rotatebox{90}{vIoU@0.5} \\
    \hline\hline
    TubeDETR~\cite{TubeDETR} & 32.5  & 20.8  & 25.7  & 8.7   & 13.6  & 6.4   & 7.2   & 2.9   & 7.8   & 3.9   & 0.7   & 0.1 \\
    STCAT~\cite{STCAT} & 36.1  & 23.2  & 34.4  & 10.4  & 14.2  & 8.4   & 3.0     & 0.1   & 7.8   & 4.4   & 0.3   & 0.0 \\
    CG-STVG~\cite{gu2024context} & 37.2  & 24.3  & 32.6  & 10.9  & 14.2  & 8.7   & 3.2   & 0.3   & 8.1   & 4.7   & 0.3   & 0.0 \\
    TA-STVG~\cite{gu2025knowing} & 38.4  & 25.2  & 35.5  & 12.1  & 13.9  & 8.5   & 3.3   & 0.2   & 7.7   & 4.5   & 0.3   & 0.0 \\
    \hline
    Baseline (ours) & 30.1  & 19.7  & 25.5  & 8.3   & 16.2  & 10.7  & 10.5  & 4.5   & 9.2   & 5.3   & 4.5   & 1.1 \\
    ART-STVG (ours) & 39.1  & 26.1  & 36.8  & 17.6  & 23.0    & 15.3  & 20.1  & 9.5   & 15.0  & 10.0   & 11.4  & 4.7 
    \\
    \hdashline
    $\Delta$ over Baseline & \improve{+9.0}  & \improve{+6.4}  & \improve{+11.3}  & \improve{+9.3}  & \improve{+6.8}   & \improve{+4.6}  & \improve{+9.6}  & \improve{+5.0}   & \improve{+5.8}  & \improve{+4.7}   & \improve{+6.9}  & \improve{+3.6} \\
    \Xhline{1.2pt}
    \end{tabular}}
  \label{tab:long-form}
  \vspace{-10pt}
\end{table}

\subsection{Comparison on Long-Form STVG }

We compare ART-STVG with other methods on the extended LF-STVG benchmarks with different average video lengths. Please \textbf{\emph{note}} that all methods including our ART-STVG are trained \emph{exclusively} on the HCSTVG-v2 training set (\ie, the average length of training videos is 20 seconds) for fair comparison.

Tab.~\ref{tab:long-form} reports the results. As shown in Tab.~\ref{tab:long-form}, our ART-STVG significantly outperforms existing STVG approaches in all metrics on all three benchmarks, showing the superiority of our method in grounding target in long-term videos. Specifically, ART-STVG outperforms TA-STVG by achieving improvements in m\_tIoU/m\_vIoU scores with 0.7\%/0.9\%, 9.1\%/6.8\%, and 7.3\%/5.5\% gains on three different video lengths, respectively. In addition, compared to our baseline, which has a similar architecture to ART-STVG but \emph{without} memory and memory selection modules (please kindly check its architecture in \emph{\textbf{supplementary material}}), ART-STVG achieves remarkable improvements in m\_tIoU/m\_vIoU with 9.0\%/6.4\%, 6.8\%/4.6\%, and 5.8\%/4.7\% gains on the three benchmarks, evidencing the importance of selective memories for LF-STVG.

\begin{table}[!t]
\setlength{\tabcolsep}{2pt}
	\centering
	\begin{minipage}{.49\textwidth}
        \centering
	\renewcommand{\arraystretch}{1.05}
        \caption{Ablation studies of the selective temporal memory in the temporal decoder. TM: Temporal Memory; TMS: Temporal Memory Selection.} \vspace{-2mm}
	\scalebox{0.75}{
		\begin{tabular}{ccccccc}
		\specialrule{1.5pt}{0pt}{0pt} 
         &
        TM & TMS &  m\_tIoU & m\_vIoT & vIoU@0.3 & vIoU@0.5 \\
			\hline
			\hline
    	  \ding{182} & - & - & 16.7 & 11.1 & 11.9  & 4.7 \\
             \ding{183} & \checkmark & - & 9.6 & 6.2 & 4.7 & 1.5 \\
             \rowcolor{cyan!10} 
             \ding{184} & \checkmark & \checkmark & \textbf{23.0} & \textbf{15.3} & \textbf{20.1} & \textbf{9.5} \\
			\specialrule{1.5pt}{0pt}{0pt}
	\end{tabular}}
        \label{tab:stm}
\end{minipage}%
\hspace{0.2em}
\begin{minipage}{.49\textwidth}
        \centering
        \caption{Ablation studies of the selective spatial memory in the spatial decoder. SM: Spatial Memory; SMS: Spatial Memory Selection.} \vspace{-2mm}
	\renewcommand{\arraystretch}{1.05}
	\scalebox{0.75}{
		\begin{tabular}{ccccccc}
			\specialrule{1.5pt}{0pt}{0pt} 
         &
         SM &  SMS & m\_tIoU & m\_vIoT & vIoU@0.3 & vIoU@0.5 \\
		\hline\hline
    	  \ding{182} & - & - & 21.3 & 13.9 & 16.4  & 8.0 \\
             \ding{183} & \checkmark & - & 22.1 & 14.2 & 17.0 & 9.0 \\
             \rowcolor{cyan!10} 
             \ding{184} & \checkmark & \checkmark & \textbf{23.0} & \textbf{15.3} & \textbf{20.1} & \textbf{9.5} \\
			\specialrule{1.5pt}{0pt}{0pt}
	\end{tabular}}
        \label{tab:ssm}
	\end{minipage}
\vspace{-5pt}
\end{table}

\begin{table}[!t]
\setlength{\tabcolsep}{2.5pt}
	\centering
	\begin{minipage}{.49\textwidth}
        \centering
    \caption{\small{Ablation studies of different designs for decoders in ART-STVG.}} \vspace{-2mm}
    \label{tab:connect}
    \scalebox{0.74}{
    \begin{tabular}{cccccc}
        \specialrule{1.5pt}{0pt}{0pt}  
         & Design & m\_tIoU & m\_vIoU & vIoU@0.3 &  vIoU@0.5 \\
        \hline
        \hline
         \ding{182} & Parallel & 21.5 & 13.9 & 17.3 & 8.2 \\
         \rowcolor{cyan!10} 
         \ding{183} & Cascaded & \textbf{23.0} & \textbf{15.3} & \textbf{20.1} & \textbf{9.5} \\
        \specialrule{1.5pt}{0pt}{0pt}
    \end{tabular}}
\end{minipage}
\hfill
\begin{minipage}{.49\textwidth}

        \centering
    \caption{Ablation studies on  $N_s$.}\vspace{-2mm}
    \label{tab:ns}
    \scalebox{0.75}{
    \begin{tabular}{cccccc}
        \specialrule{1.5pt}{0pt}{0pt} 
         
         & & m\_tIoU & m\_vIoU & vIoU@0.3 & vIoU@0.5 \\
        \hline
        \hline
         \ding{182} & $N_s=16$ & 22.7 & 15.0 & 18.4 & 9.2 \\
         \rowcolor{cyan!10} 
         \ding{183} & $N_s=32$ & \textbf{23.0} & \textbf{15.3} & \textbf{20.1} & \textbf{9.5}  \\
         \ding{184} & $N_s=48$ & 22.5 & 14.7 & 18.2 & 9.1 \\
        \specialrule{1.5pt}{0pt}{0pt}
    \end{tabular}}

	\end{minipage}
 \vspace{-5pt}
\end{table}

\subsection{Ablation Study}
To better understand ART-STVG, we further study its different designs by conducting ablation experiments on LF-STVG-3min with in-depth analysis. Our final configuration is highlighted in {\color{cyan!75} cyan}.

\vspace{0.3em}
\noindent
\textbf{Impact of selective temporal memory.} 
We set up a temporal memory bank in the temporal decoder to store temporal contextual information of target event and use this temporal memory to assist in locating the start and end of event related to the target. To verify its effectiveness for LF-STVG, we conduct an ablation study in Tab.~\ref{tab:stm}. As in Tab.~\ref{tab:stm}, without temporal memory, our method achieves a m\_tIoU score of 16.7\% (\ding{182}). When incorporating all temporal memories from the memory bank, the m\_tIoU score is decrease to 9.6\% (\ding{182} \emph{v.s.} \ding{183}). This is because the long-term video often contains multiple events, and using all temporal memories can introduce a lot of irrelevant information. With the help of memory selection, the m\_tIoU score is improved to 23.0\% with 13.4\% gains (\ding{183} \emph{v.s.} \ding{184}). These experimental results show that our selective temporal memory can effectively enhance ART-STVG for improving LF-STVG in long videos. 

\vspace{0.3em}
\noindent
\textbf{Impact of selective spatial memory.} 
Similar to the temporal decoder, we adopt a spatial memory bank in the spatial decoder to learn contextual information about the target for locating its spatial position. We conduct an ablation in Tab.~\ref{tab:ssm}. From Tab.~\ref{tab:ssm}, we observe that integrating all spatial memories can improve the m\_tIoU score to 22.1\% with 0.8\% gains (\ding{182} \emph{v.s.} \ding{183}), and applying the memory selection strategy can further enhance the m\_tIoU score to 23.0\% with 0.9\% gains (\ding{183} \emph{v.s.}~\ding{184}), validating the importance of selective memory.

\vspace{0.3em}
\noindent
\textbf{Impact of design for spatial and temporal decoders.} 
Unlike existing STVG approaches that parallelize spatial and temporal decoders, we introduce a novel cascaded spatio-temporal design for LF-STVG.  A key challenge of LF-STVG is that, as video length grows, the difficulty of temporal localization increases. To handle this, we cascade spatial and temporal decoders in ART-STVG, which allows the use of spatial localization information to assist temporal localization. We conduct an ablation in Tab.~\ref{tab:connect}. We can see that, cascading spatial and temporal decoders outperforms the parallel design, with improvements of 1.5\% and 1.4\% scores on m\_tIou and m\_vIoU (\ding{182} \emph{v.s.} \ding{183}), respectively.

\vspace{0.3em}
\noindent
\textbf{Impact of the number of selective spatial memories.} 
In the spatial decoder, we utilize $N_s$ to control the number of selective spatial memories. To explore the impact of $N_s$, we conducted the ablation experiment in Tab.~\ref{tab:ns}. We can see that when $N_s$ is 32, the performance of the model is the best (\ding{183}).

\setlength{\columnsep}{8pt}
\setlength\intextsep{0pt}
\begin{wraptable}{r}{0.5\linewidth}
\setlength{\tabcolsep}{1pt}
	\centering
        \caption{Comparison of ART-STVG with existing methods on LF-STVG using the HCSTVG-v2 validate set. The best result is highlighted in the \textbf{bold} font.} \vspace{2mm}
	\renewcommand{\arraystretch}{1.0}
	\scalebox{0.68}{
    \begin{tabular}{rcccc}	
        \specialrule{1.5pt}{0pt}{0pt} 
         Methods & m\_tIoU & m\_vIoU & vIoU@0.3 &  vIoU@0.5 \\
        \hline
        \hline
        TubeDETR ~\cite{TubeDETR} & 20.8 & 11.5 & 9.8 & 3.9  \\
        STCAT~\cite{STCAT}  & 21.0 & 12.2 & 7.4 & 0.6 \\
        CG-STVG ~\cite{cgstvg} & 20.5 & 12.0 & 8.0 & 1.0 \\ 
        TA-STVG ~\cite{gu2025knowing} & 20.7 & 11.8 & 7.7 & 0.5 \\ \hline
        ART-STVG (ours) & \textbf{28.3} & \textbf{18.8} & \textbf{27.0} & \textbf{11.9} \\
        \specialrule{1.5pt}{0pt}{0pt}
    \end{tabular}}
        \label{tab:longtrain}
 \vspace{6pt}
\end{wraptable}
\vspace{0.3em}
\noindent
\textbf{Impact of training video length.} 
To investigate the impact of training videos of different lengths, we extend HCSTVG-v2 training set to 40 seconds and use it to train both existing methods and ART-STVG. As shown in Tab.~\ref{tab:longtrain}, we can observe that all methods show improvements when trained on 40-second videos compared to 20-second videos (Tab.~\ref{tab:longtrain} \emph{v.s.}  Tab.~\ref{tab:long-form} on LF-STVG-3min).
This shows that training with longer videos enhances target localization in long-term videos, yet results in significantly increasing training costs.
More importantly, our method still achieves the best performance on all metrics, with improvements of 7.6\% score on m\_tIoU and 7.0\% score on m\_vIoU compared to TA-STVG.

\subsection{Comparison on Short-Form STVG}

% \setlength{\columnsep}{8pt}
% \setlength\intextsep{0pt}
% \begin{wraptable}{r}{0.5\linewidth}
% \vspace{-12pt}
% \setlength{\tabcolsep}{1pt}
% 	\centering
%         \caption{\small{Comparison of ART-STVG with existing methods on LF-STVG using the HCSTVG-v2 validate set. The best result is highlighted in the \textbf{bold} font.}} \vspace{2mm}
% 	\renewcommand{\arraystretch}{1.0}
% 	\scalebox{0.68}{
% 		\begin{tabular}{rcccc}
% 			\specialrule{1.5pt}{0pt}{0pt}
% 			 Methods & m\_tIoU & m\_vIoU & vIoU@0.3 &  vIoU@0.5 \\
% 			\hline\hline
% 		2D-Tan~\cite{2d-tan}  & - & 30.4 &  50.4 & 18.8  \\
% 		MMN~\cite{mmn} & - & 30.3 & 49.0 & 25.6 \\
% 		TubeDETR~\cite{TubeDETR} & 53.9 & 36.4 & 58.8 & 30.6 \\
%             STCAT~\cite{STCAT} & 56.6 & 36.9 & 60.3 & 33.6 \\
% 		STVGFormer~\cite{csdvl} & 58.1 & 38.7 & 65.5 & 33.8 \\
%             CG-STVG~\cite{cgstvg} & 60.0 & 39.5 & 64.5 & 36.3 \\
%             TA-STVG~\cite{gu2025knowing} & \textbf{60.4} & \textbf{40.2} & \textbf{65.8} & \textbf{36.7} \\   
%             \hline
%             Baseline (ours) & 46.2 & 29.9 & 45.0 & 21.1 \\
%             ART-STVG (ours) & 59.2 & 39.2 & 64.4 & 33.2 \\
% 			\specialrule{1.5pt}{0pt}{0pt}
% 	\end{tabular}}
%         \label{tab:short-form}
%  \vspace{6pt}
% \end{wraptable}
We further evaluate ART-STVG on SF-STVG in Tab.~\ref{tab:short-form} on HCSTVG-v2 validation set. As in Tab.~\ref{tab:short-form}, our method shows competitive results to current STVG methods on short-term videos. Current methods use non-autoregressive structures that process video frames in parallel to capture inter-frame relationships, and are specially designed for target localization in short-term videos. Despite this, our ART-STVG, adopting an autoregressive structure, outperforms most existing methods, falling only behind the latest state-of-the-art TA-STVG~\cite{gu2025knowing} by 1.2\% in m\_tIoU and 1.0\% in m\_vIoU. Moreover, our method shows clear gains compared to the baseline without memory.

\begin{table}[!t]
    \centering
    \setlength{\tabcolsep}{12pt} % 恢复或调整列间距（原1pt在非悬浮下可能过窄）
    \caption{\small{Comparison of ART-STVG with existing methods on LF-STVG using the HCSTVG-v2 validate set. The best result is highlighted in the \textbf{bold} font.}}
    \label{tab:short-form}
    \vspace{2mm}
    \renewcommand{\arraystretch}{1.0}
    \scalebox{0.95} { % 悬浮时为了挤空间用了0.68，正常表格可以适当放大（如0.9或直接去掉scalebox）
        \begin{tabular}{rcccc}
            \specialrule{1.5pt}{0pt}{0pt}
            Methods & m\_tIoU & m\_vIoU & vIoU@0.3 &  vIoU@0.5 \\
            \hline\hline
            2D-Tan~\cite{2d-tan}  & - & 30.4 &  50.4 & 18.8  \\
            MMN~\cite{mmn} & - & 30.3 & 49.0 & 25.6 \\
            TubeDETR~\cite{TubeDETR} & 53.9 & 36.4 & 58.8 & 30.6 \\
            STCAT~\cite{STCAT} & 56.6 & 36.9 & 60.3 & 33.6 \\
            STVGFormer~\cite{csdvl} & 58.1 & 38.7 & 65.5 & 33.8 \\
            CG-STVG~\cite{cgstvg} & 60.0 & 39.5 & 64.5 & 36.3 \\
            TA-STVG~\cite{gu2025knowing} & \textbf{60.4} & \textbf{40.2} & \textbf{65.8} & \textbf{36.7} \\
            \hline
            Baseline (ours) & 46.2 & 29.9 & 45.0 & 21.1 \\
            ART-STVG (ours) & 59.2 & 39.2 & 64.4 & 33.2 \\
            \specialrule{1.5pt}{0pt}{0pt}
        \end{tabular}
    }
\end{table}

\subsection{Qualitative Results} 

To qualitatively validate our method on LF-STVG, we present grounding results and comparisons with the baseline method without memory on the LF-STVG-3min benchmark. From Fig.~\ref{fig:qua}, we observe that the baseline without memory often locates inconsistent targets across different video frames, such as in the third and fifth examples. In contrast, our method locates the target more consistently and accurately, demonstrating the effectiveness of our approach.

\begin{figure}[!t]
    \centering
    \includegraphics[width=0.95\linewidth]{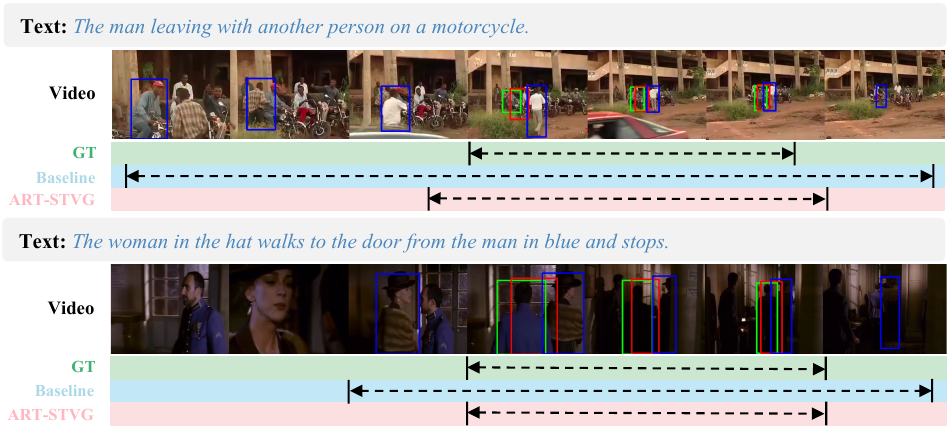}\vspace{-2mm}
    \caption{Qualitative comparison of our ART-STVG and baseline.}
    \label{fig:qua}\vspace{-5mm}
\end{figure}

\subsection{Analysis of Efficiency}

\setlength{\columnsep}{8pt}
\setlength\intextsep{0pt}
\begin{wraptable}{r}{0.55\linewidth}
\vspace{-10pt}
\setlength{\tabcolsep}{5pt}
\centering
\renewcommand{\arraystretch}{1}
\caption{Comparison on the model efficacy and complexity on
a single A100 GPU.}\vspace{2mm}
\scalebox{0.68}{
    \begin{tabular}{rcccc}
    \hline 
& \multicolumn{2}{c}{Params} & \multicolumn{2}{c}{Inference} \\  
\multirow{-2}{*}{Methods} & Trainable & Total & Time & GPU Mem \\
	\hline
		\hline
             TubeDETR~\cite{TubeDETR} & $185$ M & $185$ M & $0.23$ s & $22.1$ G  \\
             STCAT~\cite{jin2022embracing} & $207$ M & $207$ M &  $0.47$ s & $23.6$ G \\ 
			CG-STVG~\cite{cgstvg} & $203$ M & $231$ M & $0.71$ s & $25.9$ G \\
            TA-STVG~\cite{gu2025knowing} & $206$ M & $234$ M & $0.69$ s & $25.1$ G \\
            \hline
			ART-STVG (ours) & $207$ M & $235$ M & $1.09$ s & $7.9$ G \\
			\hline
	\end{tabular}}
	\label{complex}\vspace{5pt}
\end{wraptable}
To analyze the efficacy and complexity, we report the efficiency of our model and its comparison with other methods in Tab.~\ref{complex}. As shown in Tab.~\ref{complex}, our model size is similar to other methods. Although our inference time (for 64 images, aligned to SF-STVG methods) is longer due to autoregressive processing, at $1.09$ seconds, compared to the inference times of STCAT, CG-STVG, and TA-STVG, which are  $0.47$, $0.71$ and $0.69$ seconds respectively, our GPU memory usage is much lower than other methods. Specifically, our GPU memory usage is $7.9$G, while other methods such as TA-STVG and CG-STVG have GPU memory usages of $25.1$G and $25.9$G, respectively. Therefore, our method is more suitable for handling long videos.

Due to limited space, we show additional results, analyzes, and discussions in the \emph{\textbf{supplementary material}}.
\vspace{-0.1em}

\section{Conclusion}
In this work, we explore the LF-STVG problem, and propose ART-STVG to handle long-term videos. The crux of ART-STVG lies in the simple but effective use of selective memories, which are applied to decoders to leverage crucial spatio-temporal contextual information for target localization, greatly improving performance.
Additionally, our proposed cascaded spatio-temporal decoder design effectively leverages spatial localization to assist temporal localization in long-term videos. To validate the effectiveness of ART-STVG, we conduct extensive
experiments on LF-STVG benchmarks. The results show that ART-STVG significantly outperforms other methods. In addition, our method achieves comparable performance on SF-STVG benchmarks, demonstrating its generality. We hope that this work can inspire more future research on LF-STVG.

\vspace{0.5em}
\noindent
\textbf{Acknowledgment.} BF, YH, CH, and HF are not supported by any funds this this work.

\vspace{2em}

\noindent
\textbf{\Large Supplementary Material}

\vspace{0.5em}
\noindent
For a better understanding of this work, we offer additional details, analysis, and results as follows:

\vspace{-0.3em}
\begin{itemize}%[leftmargin=1.5em]

   \item \textbf{S1 Design of Baseline} \\
   In this section, we introduce the architecture of the baseline method.
   
   \item \textbf{S2 Analysis of Failure Cases} \\
   In this section, we discuss failure cases of our proposed method.
   
   \item \textbf{S3 Qualitative Results} \\
   In this section, we show more qualitative results of our method on LF-STVG and comparison to the baseline method.

   \item \textbf{S4 Ablation Results and aditionanl analysis} \\
   In this section, we present additional analyses and experiments.

   \item \textbf{S5 Limitation and Broader Impact} \\
   In this section, we discuss the limitation and broader impact of our method.
   
\end{itemize}

\begin{figure}[b]
    \centering
\includegraphics[width=1.\linewidth]{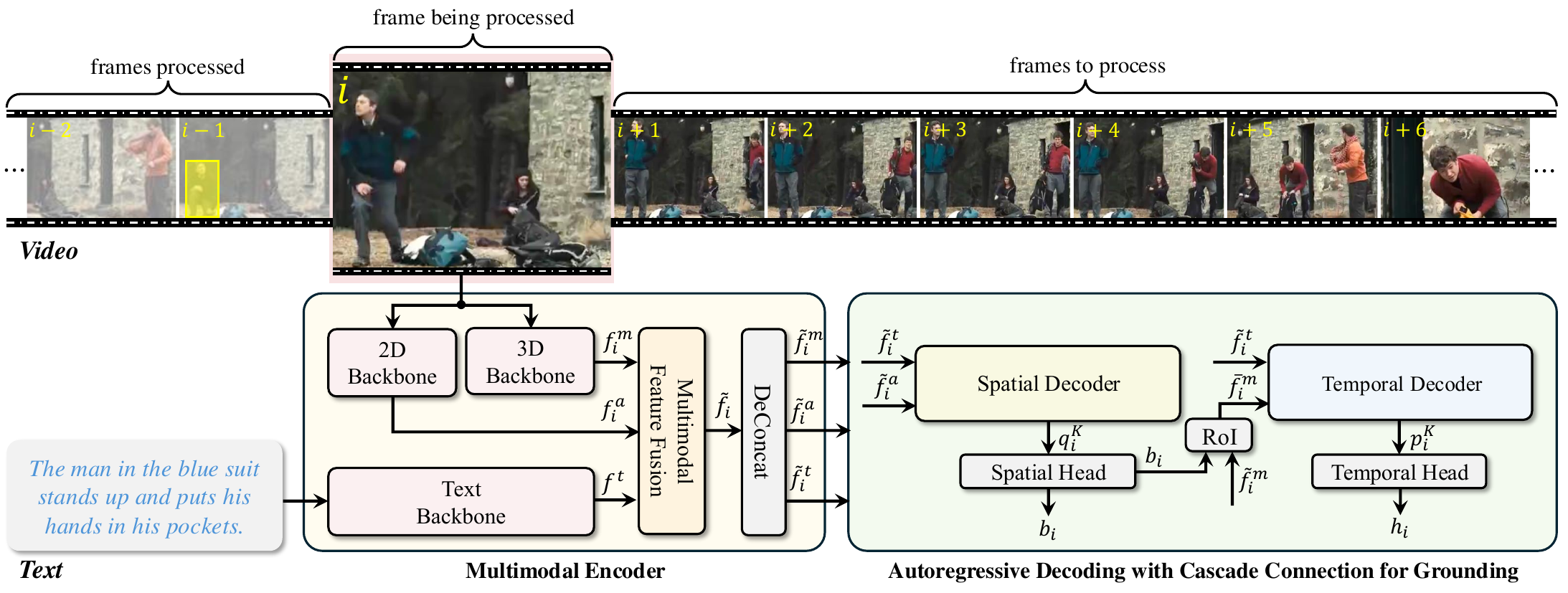}%\vspace{-2mm}
\caption{Architecture of baseline, containing an encoder and an autoregressive decoder but \emph{without} memories and memory-related modules.}
    \label{fig:base}
\end{figure}

\section*{S1 \; Design of Baseline}
The baseline method mentioned in the main text shares a similar architecture with ART-STVG but does \emph{not} contain the (spatial and temporal) memories and memory-related modules. Its detailed architecture is shown in Fig.~\ref{fig:base}. Specifically, the baseline framework comprises two core components: a multimodal encoder for extracting and fusing cross-modal features, followed by cascaded spatial and temporal decoders that perform autoregressive decoding to progressively localize the target. Such architecture makes it suitable for long video processing. However, without memory, its performance is inferior to our ART-STVG (please see the comparison of our ART-STVG and the baseline in the main text), which evidences the necessity and importance of our memory design for improving LF-STVG performance.

\section*{S2 \; Analysis of Failure Cases}

\setlength{\columnsep}{10pt}%
\setlength\intextsep{0pt}
\begin{wrapfigure}{r}{0.55\textwidth}
\centering
%\vspace{-5mm}
\includegraphics[width=0.55\textwidth]{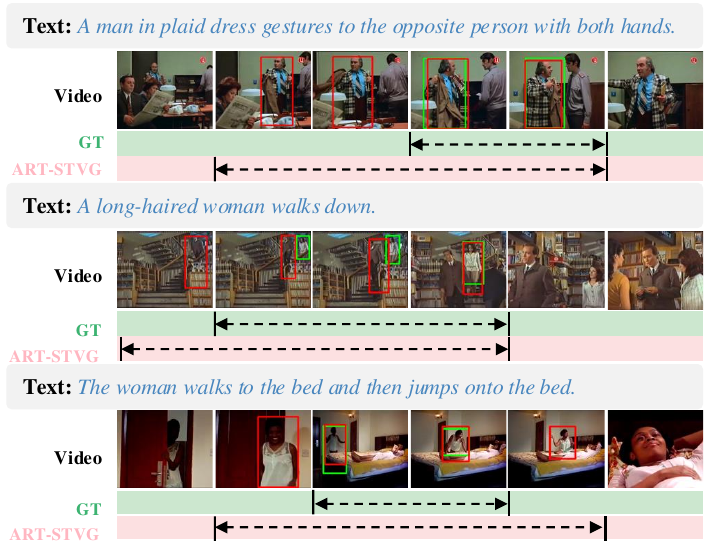}
\caption{Failure cases. The red box in the figures indicates the results of \textcolor{lightpink}{ART-STVG}, while the green box indicates the \textcolor{mediumgreen}{ground truth (GT)}.}
%\vspace{3mm}
\label{fig:fc}
\end{wrapfigure}
Despite showing promising performance of LF-STVG, our method may fail in some complex scenes: \emph{(i) indistinct event boundaries}. Detection of event boundaries in long videos is crucial for temporal memory selection in ART-STVG. When event boundaries are ambiguous in videos, their detection may be compromised, which results in inaccurate temporal memory selection and hence degrades temporal localization in STVG; for example, in Fig.~\ref{fig:fc}-top, the start time of the target event is ambiguous. \emph{(ii) highly distracting background objects}. When there exist highly distracting background targets (that have similar appearance and action with the foreground target) in videos, our method may drift to background targets, leading to spatial localization failure. In Fig.~\ref{fig:fc}-middle, there are two people with similar actions. \emph{(iii) Extremely short target events.} When the target event lasts for a very short duration within a long video, the presence of a large amount of redundant information makes grounding more difficult. For example, in Fig.~\ref{fig:fc}-bottom, the target event lasts only 3 seconds, while the entire video is 300 seconds long, making localization hard. To handle the above cases, we will explore fine-grained target and event cues in videos for improvements in our future work.

\section*{S3 \; Analysis of Qualitative Results}

To qualitatively validate our method on LF-STVG, we present grounding results and comparisons with the baseline method without memory on the LF-STVG-3min benchmark. From Fig.~\ref{fig:qua}, we observe that the baseline without memory often locates inconsistent targets across different video frames, such as in the third and fifth examples. In contrast, our method locates the target more consistently and accurately, demonstrating the effectiveness of our approach.

\begin{figure}[htb]
    \centering
    \includegraphics[width=1.0\linewidth]{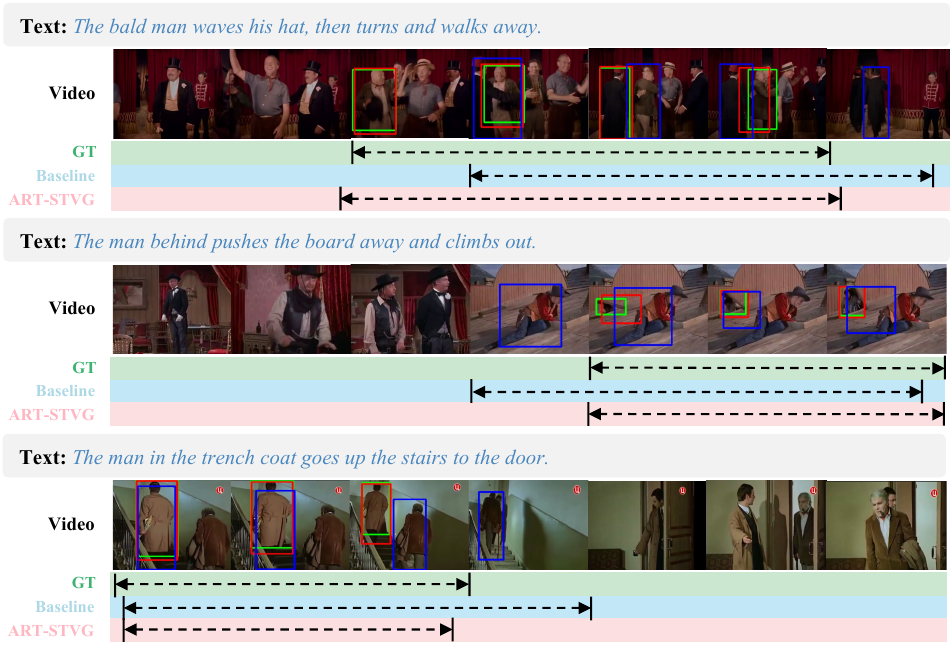}%\vspace{-2mm}
    \caption{Qualitative comparison of our ART-STVG and the baseline.}
    \label{fig:qua}%\vspace{-4mm}
\end{figure}

\section*{S4 Ablation Results and aditionanl analysis}

\noindent
\subsection*{4.1 Ablation of threshold $\theta_t$ in temporal memory selection} 
To investigate the impact of the temporal similarity threshold $\theta_t$ on our temporal memory selection mechanism, we conduct an ablation study by varying $\theta_t$ among $\{0.7, 0.8, 0.9\}$ across three datasets of different video lengths: LF-STVG-1min, LF-STVG-3min, and LF-STVG-5min. As shown in Tab. A, setting $\theta_t$ to $0.8$ consistently achieves the best performance across all datasets. When the threshold is set lower at $0.7$, the performance drops slightly (e.g., to $38.7\%$/$25.5\%$ on LF-STVG-1min) because a looser criterion introduces redundant, noisy temporal contexts that distract the model's attention. Conversely, raising the threshold to a stricter value of $0.9$ leads to a more pronounced performance degradation (e.g., dropping to $14.4\%$/$9.5\%$ on LF-STVG-5min) as it over-filters valuable historical memories, depriving the model of necessary long-term temporal dependencies. These results demonstrate that $\theta_t = 0.8$ strikes an optimal balance between noise suppression and context preservation, exhibiting strong robustness across various video durations.

\begin{table}[htb]
	\centering
	\setlength{\tabcolsep}{6pt}
	\renewcommand{\arraystretch}{1.1}
    \caption{Ablation study on $\theta_t$.}
	\scalebox{0.8}{
    \begin{tabular}{c|cc|cc|cc}
    \Xhline{1pt}
    \rowcolor{mygray}
    \cellcolor{mygray} & \multicolumn{2}{c|}{ \cellcolor{mygray} LF-STVG-1min} & \multicolumn{2}{c|}{ \cellcolor{mygray}LF-STVG-3min} & \multicolumn{2}{c}{ \cellcolor{mygray}LF-STVG-5min} \\
    \rowcolor{mygray}
    \multirow{-2}{*}{\cellcolor{mygray} $\theta_t$} & {m\_tIoU} & {m\_vIoU} & {m\_tIoU} & {m\_vIoU} & {m\_tIoU} & {m\_vIoU} \\
	\hline
		\hline
			0.7 & 38.7 & 25.5 & 22.7 & 15.1 & 14.7 & 9.8 \\
			0.8 & \textbf{39.1}  & \textbf{26.1}  & \textbf{23.0}    & \textbf{15.3}  & \textbf{15.0}  & \textbf{10.0}\\
            0.9 & 38.4 & 25.3 & 22.8 & 15.0 & 14.4 & 9.5\\
			\Xhline{1pt}
	\end{tabular}}
	\label{tab:loss_weight_ablation}
\end{table}

\subsection*{4.2 Analysis of streaming input for LF-STVG}
As video length increases, existing STVG models that process all frames simultaneously face severe GPU memory limitations. For example, processing just 284 frames requires approximately 80GB of GPU memory for conventional methods like TubeDETR, STCAT and CG-STVG, which already saturates the memory capacity of an NVIDIA A100 GPU. This memory bottleneck fundamentally prevents these full-batch approaches from scaling to longer videos.

\noindent
In contrast, our ART-STVG adopts a streaming processing paradigm that processes frames sequentially rather than loading the entire video into memory at once. As shown in Fig. This streaming approach not only alleviates the GPUs memory burden but also enables practical processing of the extended video lengths present in the LF-STVG benchmark, which would be infeasible with conventional full-video processing architectures.

\begin{figure}[htbp]
\centering
\includegraphics[width=0.8\textwidth]{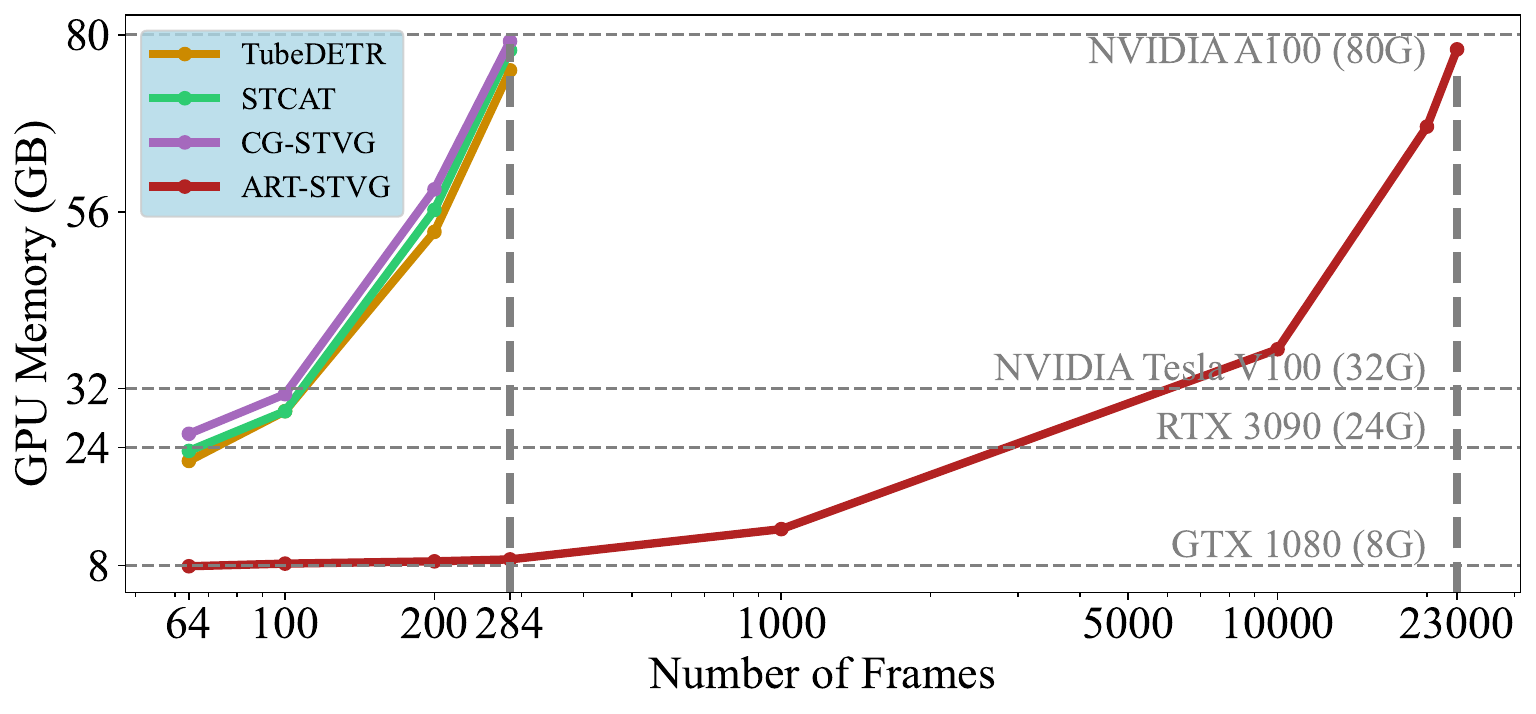}
\caption{GPU memory consumption as the video length grows. Please note: the accuracy may decrease.}
\label{fig:length}
\end{figure}

\subsection*{4.3 Latency and memory analysis for long videos.}
To demonstrate the efficiency and practicality of our proposed memory mechanism for long-form video understanding, we analyze its computational latency and memory overhead.
Our memory mechanism is extremely lightweight: the memory update and retrieval processes add only a negligible latency overhead of approximately $1\%$ to the overall inference time, ensuring high-speed streaming.
Regarding the memory footprint, each spatial or temporal memory entry is represented as a compact $1 \times 256$ float32 vector, requiring only $2$~KB of memory per frame.
Under a standard sampling rate of 3.2~fps, storing the history of an entire 60-minute video (amounting to 11,520 frames) requires only about $22.5$~MB of memory.
This exceptionally low memory footprint, combined with our streaming paradigm, makes our framework highly scalable and uniquely suited for processing extremely long-form videos on standard hardware, where conventional full-video models would inevitably fail due to out-of-memory (OOM) errors.

\section*{S5 Limitation and Broader Impact}

\noindent
\textbf{Discussion of Limitation.} As the first work to explore the LF-STVG problem, our ART-STVG shows promising performance on long-term videos and significantly outperforms existing STVG methods. Despite this, our method has two limitations. First, although ART-STVG is capable of handling long videos, yet its performance may degrade as the video becomes longer and more complex. To mitigate this issue, a possible direction is to learn more discriminative memory systems for capturing fine-grained target information. Since this is beyond our current goal of attempting the LF-STVG problem, we leave it to our future work for improving LF-STVG. Second, similar to other approaches, our method cannot operate in real-time, which may limit its applications in certain scenarios. In the future, we will study lightweight architectures for LF-STVG using techniques from model compression and quantization.

\vspace{0.5em}
\noindent
\textbf{Discussion of Broader Impact.} The proposed ART-STVG focuses on localizing the target of interest within an untrimmed video given a textual description. This technique has a wide range of crucial applications, including content-based video retrieval, video content moderation, and sports analytics. By developing ART-STVG, we expect it to can contribute positively to societal advancements and technological progress.

\bibliographystyle{splncs04}
\bibliography{main}
\end{document}